\documentclass{article}

\usepackage[nonatbib, final]{nips_2016}

\usepackage[utf8]{inputenc} %
\usepackage[T1]{fontenc}    %
\usepackage{hyperref}       %
\usepackage{url}            %
\usepackage{booktabs}       %
\usepackage{amsfonts}       %
\usepackage{nicefrac}       %
\usepackage{microtype}      %

\usepackage{footnote}
\usepackage{graphicx}
\usepackage{setspace}

\usepackage{verbatim}

\newcommand*\rot{\rotatebox{90}}

\title{
    Bridging Category-level and Instance-level Semantic Image Segmentation\thanks{This
    research was in part supported by the Data to Decisions
    Cooperative Research Centre.
    C. Shen's participation was in part
    supported by an ARC Future Fellowship (FT120100969).
    C. Shen is the corresponding author.
  }
 }

\author{
  Zifeng Wu, Chunhua Shen, Anton van den Hengel\\
  The University of Adelaide,
  SA 5005, Australia\\
  \texttt{firstname.lastname@adelaide.edu.au}\\
}

\begin{document}

\maketitle

\begin{abstract}

We propose an approach to instance-level image segmentation  that is built on top of  category-level segmentation.
    Specifically,
    for each pixel in a semantic category mask,  its corresponding instance bounding box is predicted
    using a deep fully convolutional regression network.
Thus it follows a different pipeline to the popular detect-then-segment approaches that first predict instances' bounding boxes,
which are the current state-of-the-art in instance segmentation.
We show that, by leveraging the strength of our state-of-the-art semantic segmentation models,
the proposed method can achieve comparable or even better results to detect-then-segment approaches.
We make the following contributions.
(i) First, we propose a simple yet effective approach to semantic instance segmentation.
(ii) Second, we propose an online bootstrapping method during  training,
which is critically important for achieving good performance for both semantic category segmentation  and instance-level segmentation.
(iii) As the performance of semantic category segmentation has a significant impact on
    the instance-level segmentation,
    which is the second step of our approach,
    we train fully convolutional residual networks to achieve the best semantic category segmentation accuracy.
    On the PASCAL VOC 2012 dataset, we obtain the currently best mean intersection-over-union score of 79.1\%.
(iv) We also achieve state-of-the-art results for instance-level segmentation.

\end{abstract}

\section{Introduction}

Semantic category-level image segmentation amounts to predicting the category of individual pixels in an image,
which has been one of the most active topics in the field of image understanding and computer vision for a long time.
Most of the recently proposed approaches to this task are based on deep convolutional networks.
Particularly, the fully convolutional network (FCN)~\cite{FCN.CVPR.2015.Long} is
efficient and at the same time has achieved the state-of-the-art performance.
By reusing the computed feature maps for an image,
FCN avoids redundant re-computation for classifying individual pixels in the image.
FCN has become the \emph{de facto} approach to dense prediction,
and many methods have been proposed to further improve this framework,
e.g., the DeepLab~\cite{DeepLab.ICLR.2015.Chen} and the Adelaide-Context model \cite{AdelaideContext.2016.Lin}.
One key reason for the success of these methods is that they are based on
rich features learned from the very large ImageNet~\cite{ILSVRC2015} dataset,
often in the form of a 16-layer VGGNet~\cite{VGGNet.2014.Simonyan}.
However, currently, there exist much improved models for image classification, e.g.,
the ResNet~\cite{ResNet.CVPR.2016.He,ResNet2.2016.He}.

Semantic instance-level segmentation aims to identify the individual instances of different semantic categories,
i.e., simultaneous object detection and segmentation.
Instance segmentation is supposed to be  only one more step beyond semantic segmentation.
However, most recently proposed approaches~\cite{SDS.ECCV.2014.Hariharan,HyperColumn.CVPR.2015.Hariharan,MNC.CVPR.2016.Dai} to instance segmentation are based on bounding-box detection methods.
Generally speaking, these methods follow the pipeline of a typical object detection method,
e.g., Fast R-CNN~\cite{FastRCNN.ICCV.2015.Girshick},
to locate instances of semantic objects with bounding-boxes {\em as the first step},
and then predict binary object masks within these boxes via background-foreground segmentation.
In spite of the recent fast development of powerful semantic-category segmentation methods,
there is little work in the literature towards developing an instance segmentation method on top of, and fully exploiting the power of such
methods.
One notable work is the proposal-free network~(PFN)~\cite{ProposalFree.2015.Liang},
which for all pixels predicts semantic scores and instance bounding-boxes, as well as category-wise numbers of instances per image,
based on which a clustering algorithm is applied for post-processing to locate instances.
As pointed out by Uhrig~et~al.~\cite{CityscapesInstance.2016.Uhrig},
PFN has a complex architecture with many interleaved building blocks, which makes training very complex.
More importantly, its performance seemingly depends critically on the correct prediction of instance numbers,
which is sometimes infeasible. Especially, in a complex scene usually there are many small instances, and
the number of training samples per number of instances can be very small,
leading to mistakes in their estimation. Consequently, the available cues for
clustering may significantly deviate from the estimated number of instances.
This might be one of the reasons why
their performance reported in \cite{ProposalFree.2015.Liang} is considerably poorer than most recent results.
On the other hand, the concurrent work to ours by Uhrig~et~al.~\cite{CityscapesInstance.2016.Uhrig} relies on elaborately designed template matching to detect instances,
which is very case-specific. Furthermore, good performance in~\cite{CityscapesInstance.2016.Uhrig} heavily
relies on  accurate depth estimation, which only works well   for indoor/outdoor scene images.
Based on the above considerations,
here we attempt to develop a concise and generic yet accurate method of this kind,
and show the great potential of the less explored route.

In summary, we highlight the main contributions of this work as follows:
\vspace{-2.0mm}
\begin{itemize}
\item
We propose a new approach to instance-level segmentation, in which semantic score maps are transformed into Hough-like maps.
We can easily detect the instances of different semantic objects from these transformed maps.
\vspace{-1.0mm}
\item
We propose an online bootstrapping method for training,
and show that it is critically important in achieving the best performance both for semantic and instance segmentation.
\vspace{-1.0mm}
\item
We extensively evaluate different variations of a fully convolutional residual network~(FCRN) so as to find the best configuration,
including the number of layers, the resolution of feature maps, and the size of field-of-view.
\vspace{-1.0mm}
\item
We introduce dropout regularization to some of the residual blocks in an FCRN, replace the top-most linear classifier with a multi-layer non-linear one, and adopt the multi-view testing technique,
by which we further improve the performance.
Our method achieves the currently best results on the PASCAL VOC 2012 dataset in terms of semantic segmentation.
Our mean intersection-over-union score reaches 79.1\% even though we use only the augmented PASCAL VOC training data,
which is a new record\footnote{\url{http://host.robots.ox.ac.uk:8080/anonymous/MZVIPW.html}}.
\vspace{-1.0mm}
\item
We achieve \emph{on par} or even better results in terms of instance segmentation on the PASCAL VOC 2012 dataset, compared with the previous best performers.
In particular, we significantly improve the mean region average precision at an overlap of 0.7 by 5.1\%, from 41.5\% to 46.6\%.
We empirically show that, as the accuracy of semantic segmentation improves, our method has the potential to improve by a remarkable margin.
\end{itemize}

\textbf{Related work}
Next  we briefly review the most recent developments within four topics, which are closely related to this paper.

\textbf{Very deep convolutional networks.}
The recent boom in deep convolution networks originated when Krizhevsky~et~al.~\cite{CNN.NIPS.2012.Krizhevsky} won the first place in the ILSVRC 2012 competition~\cite{ILSVRC2015} with the 8-layer AlexNet.
In the next year, `Clarifai'~\cite{ILSVRC2015} still had the same number of layers.
However, in 2014, the VGGNets~\cite{VGGNet.2014.Simonyan} were composed of up to nineteen layers,
while the even deeper 22-layer GoogLeNet~\cite{GoogLeNet.2014.Szegedy} won the competition~\cite{ILSVRC2015}.
In 2015, the much deeper ResNets~\cite{ResNet.CVPR.2016.He} achieved the best performance~\cite{ILSVRC2015},
showing deeper networks indeed learn better features.

The most impressive part was that, by replacing the VGGNets in Fast RCNN~\cite{FastRCNN.ICCV.2015.Girshick} with their ResNets, He~et~al.~\cite{ResNet.CVPR.2016.He} won in the object detection task with a remarkable margin.
This result showed the importance of features in image understanding tasks.
The main contribution that enabled them to
train such deep networks was that they connected some of the layers with shortcuts.
These shortcuts directly passed through the signals,
and thus avoid the gradient vanishing effect,
which may be a problem for very deep plain networks.
In a more recent work, they redesigned their residual blocks to avoid over-fitting,
which enabled them to train an even deeper 200-layer residual network.
Deep ResNets can be seen as a simplified version of the highway network~\cite{Highway}.

\textbf{Fully convolutional networks for semantic segmentation.}
Long~et~al.~\cite{FCN.CVPR.2015.Long} first proposed the framework of fully convolutional networks for semantic segmentation, which was both effective and efficient.
They also enhanced the final feature maps with those from intermediate layers, which enabled their model to make finer predictions.
Chen~et~al.~\cite{DeepLab.ICLR.2015.Chen} increased the resolution of feature maps by spontaneously
removing some of the down-sampling operations and accordingly introducing kernel dilation into their networks.
They also found that a classifier composed of small kernels with a large dilation performed as well as a classifier
with large kernels, and that reducing the size of field-of-view had an adverse impact on performance.
As post-processing, they applied dense CRFs~\cite{DenseCRF.NIPS.2011.Krahenbuhl} to the network-predicted category-wise score maps for further improvement.
Zheng~et~al.~\cite{CRFasRNN.ICCV.2015.Zheng} simulated the dense CRFs with an
recurrent neural network~(RNN), which can be trained end-to-end together with the down-lying convolution layers.
Lin~et~al.~\cite{AdelaideContext.2016.Lin} jointly trained CRFs with down-lying convolution layers.
Thus they were able to capture both `patch-patch' and `patch-background'
context with CRFs, rather than just pursuing local smoothness as most of the previous methods did.

\textbf{Bounding-box detection based approaches to instance segmentation.}
Most of the best performers for instance-level segmentation in the literature can be attributed as bounding-box detection based approaches~\cite{SDS.ECCV.2014.Hariharan,HyperColumn.CVPR.2015.Hariharan,MNC.CVPR.2016.Dai}.
Usually, they were composed of two steps, i.e., first locating objects with bounding-boxes,
and second generating masks from these boxes via foreground segmentation.
However, our method in this paper is built upon semantic image segmentation.
An object is detected as a local maxima on a transformed semantic score map,
which has some connections to the generalized Hough transform based approaches to object detection and segmentation~\cite{ISM.IJCV.2007.Leibe}.

\textbf{Online bootstrapping for training deep convolutional networks.}
There are some recent works in the literature exploring sampling methods during training, which are concurrent with ours.
Loshchilov and Hutter~\cite{OHEM.ICLR.2016.Loshchilov} studied mini-batch selection in terms of image classification.
They picked hard training images from the whole training set according to their current losses, which were lazily updated once an image had been forwarded through the network being trained.
Shrivastava~et~al.~\cite{DetOHEM.CVPR.2016.Shrivastava} in a concurrent work to ours selected hard regions-of-interest (RoIs) for object detection.
They only computed the feature maps of an image once, and forwarded all RoIs of the image on top of these feature maps.
Thus they were able to find the hard RoIs with a small additional computational cost.

The method of~\cite{OHEM.ICLR.2016.Loshchilov}
is similar to ours in the sense that they both select hard training samples based on the current losses of individual data-points.
However, we only search for hard pixels within the current mini-batch, rather than the whole training set.
In this sense, the method of~\cite{DetOHEM.CVPR.2016.Shrivastava} is more similar to ours.
Nevertheless, to our knowledge, the method here is the first to propose online bootstrapping of hard
pixel samples for the problems of instance and semantic image segmentation.

\section{Proposed method}

We first introduce the proposed pipeline for instance segmentation,
then demonstrate our online bootstrapping approach,
which is one of the key components in our high-performance fully convolutional residual networks~(FCRNs) for both semantic and instance segmentation,
and finally explain how to build up our FCRNs from residual networks.

\subsection{Proposed pipeline for instance segmentation}

\begin{figure}[t!]
\centering
\includegraphics[width=0.9\linewidth,trim=0 340 0 50]{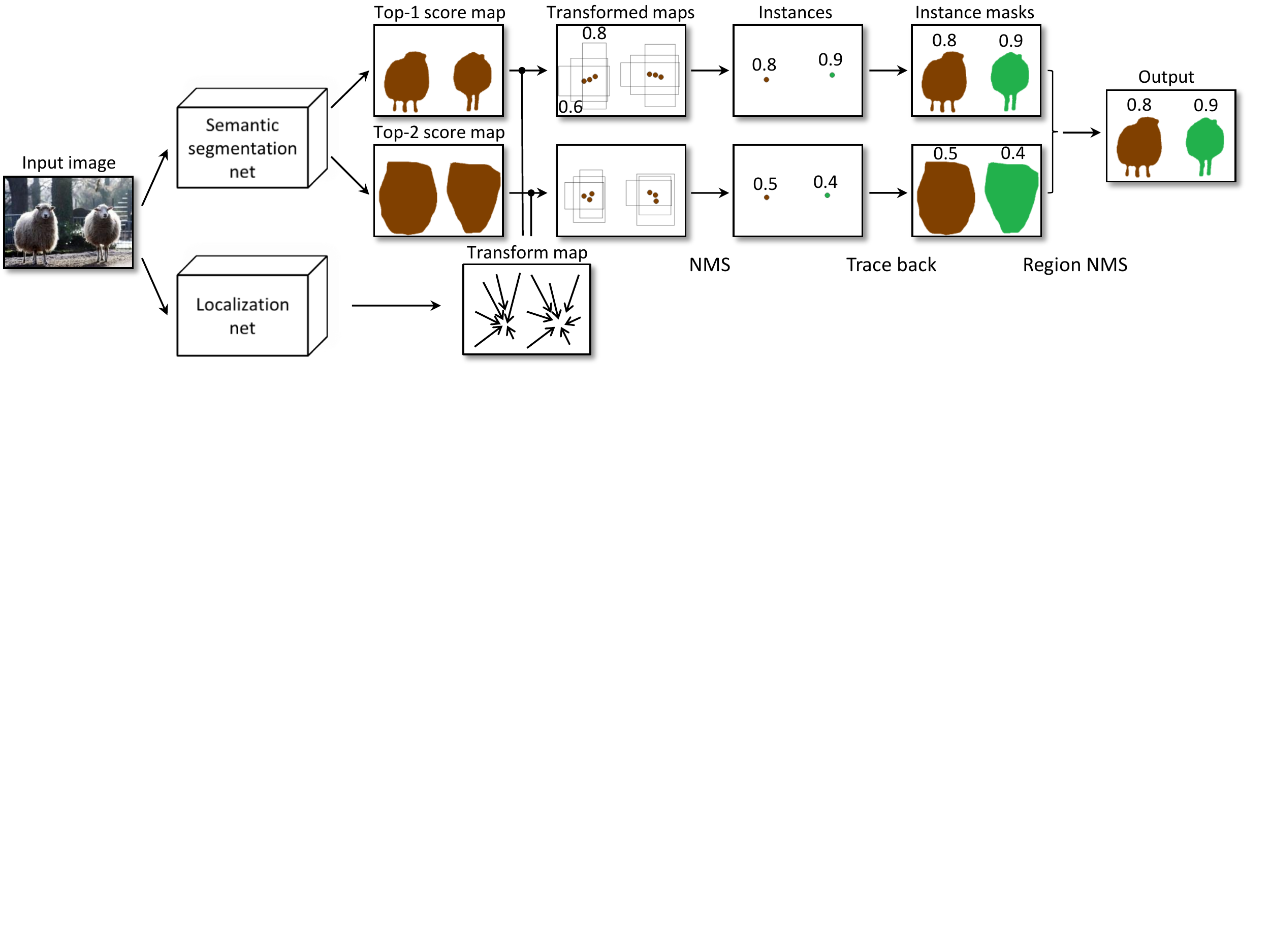}
\caption{The pipeline of our proposed approach to instance segmentation.}
\label{fig:pipeline}
\end{figure}

The pipeline of our approach is illustrated in~Fig.~\ref{fig:pipeline}.
During testing, we obtain the instance segmentation result of an image following the steps shown below.

{\bf 1}) Calculate the category-wise score maps via semantic category segmentation.
For clarity, we only depict the score maps corresponding to the `sheep' category and discard the remaining nineteen object categories in the figure.
\vspace{-1.5mm}

{\bf 2}) Calculate the category-wise transform maps via bounding-box regression.
Again, only the transform map for the `sheep' category is depicted.
\vspace{-1.5mm}

{\bf 3}) Apply the obtained transform maps to their corresponding scores maps.
\vspace{-1.5mm}

{\bf 4}) Search for local maxima on the transformed maps by non-maximal suppression~(NMS),
and keep the obtained maxima as detected instance hypotheses.
\vspace{-1.5mm}

{\bf 5}) Trace back to all of the suppressed pixels and recover a mask for each instance hypothesis.
\vspace{-1.5mm}

{\bf 6}) Generate the final instance segmentation result by region based NMS~\cite{SDS.ECCV.2014.Hariharan}.

We train the two networks separately.
The semantic segmentation net is trained with the classic Logistic regression loss,
and the localization network is trained with the smoothed $\ell_1$ loss~\cite{FastRCNN.ICCV.2015.Girshick}.
Our regression targets include the vertical and horizontal offsets from the current pixel to the bounding-box center of the instance that it belongs to,
and also the height and width of that instance.
To balance the contributions of instances in different sizes,
we re-weight their pixels in the training loss according to their heights and widths.
However, we let pixels in the same instance have the same loss weight,
since the central and peripheral pixels are of the same importance in order to reconstruct the whole instance.

To combine the outputs of both networks during testing, we start by applying the transform maps to the semantic score maps.
Note that we ignore the background and only transform the foreground pixels,
e.g., only those labeled as `sheep' in~Fig.~\ref{fig:pipeline}.
To improve the recall rate of instances, we use top-$n$ `sheep' masks.
In other words, we generate $n$ masks per category.
The top-$n$ masks are  obtained by keeping the pixels whose scores for the `sheep' category are among their top-$n$ highest scores.

The next step is to find the instances.
We need to detect the modes of transformed pixels in the 2D spatial image space.
To this end, here we resort to a simple off-the-shelf approach, although more sophisticated clustering methods
may lead to improved results.
Namely, we apply non-maximum suppression (NMS) to the bounding-boxes previously predicted by our localization net.
The pixels with locally high semantic scores will be kept as instance hypotheses.
For each of them, we view that pixel and its suppressed pixels as a cluster,
and average their semantic scores to compute the confidence of their corresponding instance hypothesis.
Note that the above process has some connections with
the generalized Hough transform based approaches to object detection and segmentation~\cite{ISM.IJCV.2007.Leibe},
and that predicting bounding-boxes is not an indispensable component but a trivial
implementation of maxima searching in our proposed method.
However, in most of the recent methods~\cite{SDS.ECCV.2014.Hariharan,HyperColumn.CVPR.2015.Hariharan,MNC.CVPR.2016.Dai},
bounding-boxes are one of the required inputs for their segmentation components to generate binary masks for individual instances.

\subsection{Online bootstrapping of hard training pixels}

When we train an FCRN, depending on the size of image crops, there may be tens of thousands of labeled pixels to predict per crop.
However, sometimes many of them can easily be discriminated from others,
especially those lying at the center of a large semantic region.
Continuing to learn from these pixels cannot improve our objective.
For this reason,
we propose an online bootstrapping method, which forces networks to
focus on hard (and so more valuable) pixels during training.

We first describe the proposed online bootstrapping in the context of semantic  category-level segmentation.
Let there be $K$ different categories $c_j$ in a label space.
For simplicity, suppose that there is only one image crop per mini-batch, and let there be $N$ pixels $a_i$ to predict in this crop.
Let $y_i$ denote the ground-truth label of pixel $a_i$, and $p_{ij}$ denote the predicted probability of pixel $a_i$ belonging to category $c_j$.
Then, the loss function can be defined as,
\begin{equation}
\ell = - \frac{1}{ \sum_{i}^{N} \sum_{j}^{K}  {\bf 1}\{y_i = j \textrm{ and } p_{ij} < t\} } ( \sum_{i}^{N} \sum_{j}^{K}
                        {\bf 1}\{y_i = j \textrm{ and } p_{ij} < t\} \log p_{ij})
\end{equation}
where $t \in (0, 1]$ is a threshold. Here
${\bf 1} \{\cdot\}$ equals one when the condition inside the brackets holds, and otherwise equals zero.
In other words, we drop pixels when they are too easy for the current model, i.e., their losses are below $t$.
However, in practice, we hope that there should be at least a reasonable number of pixels kept per mini-batch.
Otherwise, the computed gradient can become very noisy.
Hence, we  increase the threshold $t$ accordingly if our current model performs fairly well on a specific mini-batch;
and decrease if the model does not work well.

The idea to introduce online bootstrapping into our localization network is similar, i.e., to drop those too easy pixels.
However, the definition of `easy' is slightly different.
Here, we threshold the intersection-over-union~(IoU) scores between ground-truth and predicted bounding-boxes, rather than directly thresholding the regression loss.
Our intuition is straightforward.
We do not care how accurate the heights, widths or offsets are regressed, but we do care if we can get the correct instances after the following NMS on transformed score maps.
Since NMS is based on IoU scores between bounding-boxes, our choice is more natural than thresholding regression losses of the four targets respectively.

Note that this online bootstrapping approach is also capable of automatically balancing biased training data,
which is one of the significant and common problems for a pixel labeling task.
Usually, there are remarkably more background pixels than object pixels.
Sometimes, one category has many more pixels than the others.
The mechanism works as follows.
As the training process goes on, our model may become overly-learned for those majority categories.
Therefore, pixels belonging to these categories will become easy to discriminate,
which means that the training losses for these pixels will go down.
At the same time, the losses for pixels belonging to minority categories keep unchanged or even go up.
At some point, our bootstrapping approach will find them and keep the model learning from them
until their losses go down to the same level as the majority category pixels.
Similar argument was also presented in a concurrent work to ours by Shrivastava~et~al.~\cite{OHEM.ICLR.2016.Loshchilov}.

\subsection{Fully convolutional residual network}

We initialize an FCRN from the version of ResNet in~\cite{ResNet.CVPR.2016.He}.
We replace the linear classification layer with a convolution layer so as to make one prediction per spatial
location.
Besides, we also remove the 7$\times$7 pooling layer.
This layer can enlarge the field-of-view (FoV)~\cite{DeepLab.ICLR.2015.Chen} of features,
which is sometimes useful considering the fact that we human usually tell the category of a pixel by referring
to its surrounding context region.
However, this pooling layer at the same time smoothes the features.
In pixel labeling tasks, features of adjacent pixels should be distinct from each other when they respectively
belong to different categories, which may conflict with the pooling layer.
Therefore we remove this layer and let the linear convolution layer be  on top to deal with the FoV.

Up to now, the feature maps below the added linear convolution layer only has a resolution of 1/32,
which is too low to precisely discriminate individual pixels.
Long~et~al.~\cite{FCN.CVPR.2015.Long} learned extra up-sampling layers for this issue.
However, Chen~et~al.~\cite{DeepLab.ICLR.2015.Chen} reported that
the hole algorithm (or the \`{a}trous algorithm by Mallat~\cite{WaveletTour.2008.Mallat})  can be  more efficient.
Intuitively, the hole algorithm can be seen as dilating the convolution kernels before applying them to their input feature maps.
With this technique, we can build up a new network generating feature maps of any higher resolution, without changing the weights.
When there is a layer with down-sampling, we
skip the down-sampling part and increase the dilations of subsequent convolution kernels accordingly.
See DeepLab~\cite{DeepLab.ICLR.2015.Chen} for an explanation.

A sufficiently large FoV was reported to be important by Chen~et~al.~\cite{DeepLab.ICLR.2015.Chen}.
Intuitively, we need to present context information of a pixel to the top-most classification layer.
However, the features at different locations should be discriminative at the same time
so that the classifier can tell the differences between adjacent pixels which belong to different categories.
Therefore, a natural way is to let the classifier to handle the FoV,
which can be achieved by enlarging its kernel size.
Unfortunately, the required size can be so large that it can blow up the number of parameters in the classifier.
Nevertheless, we can resort to the hole algorithm again.
In other words,
we use small kernels with large dilations,
in order to realize a large FoV.

Following the above three steps, we design our baseline FCRN for semantic category-level segmentation.
Although the ResNet has shown its advantages in terms of many tasks due to much richer learned features,
we observe that this baseline FCRN is not powerful enough to beat the best algorithm for
semantic segmentation~\cite{AdelaideContext.2016.Lin},
which is based on the VGGNet~\cite{VGGNet.2014.Simonyan}.
However,
we empirically show that FCRN can achieve the best performance together with our proposed online bootstrapping and a few modifications.
We design our localization networks similarly.
We do not regress bounding-boxes for background pixels, thus there are only eighty channels in the top-most layer.

\section{Experiments}

In this section, we show that our method achieves the state-of-the-art performance both in terms of semantic
category-level and instance-level segmentation.
We implement our algorithm with the Caffe~\cite{Caffe.2014.Jia} toolkit throughout all the experiments.

\subsection{Semantic category-level segmentation results}
We first evaluate the semantic category segmentation component in our method,
which is derived based on the FCN~\cite{FCN.CVPR.2015.Long} and the ResNet~\cite{ResNet.CVPR.2016.He}.
We name it as the fully convolutional residual network~(FCRN), whose hyper-parameters are carefully evaluated, including the network depth, the resolution of feature maps, the kernel size and dilation of the top-most classifier in the network.
For evaluation, we use three popular and challenging datasets,
i.e., the PASCAL VOC 2012~\cite{PascalVoc.IJCV.2014.Everingham}, the Cityscapes~\cite{Cityscapes.CVPR.2016.Cordts}, and the \mbox{PASCAL-Context}~\cite{PascalContext.CVPR.2014.Mottaghi} dataset.
We report,
1) the pixel accuracy, which is the percentage of correctly labeled pixels on a whole test set,
2) the mean pixel accuracy, which is the mean of class-wise pixel accuracies, and
3) the mean class-wise intersection-over-union (IoU) scores.
Also note that we only show these three scores when it is possible for the individual datasets.
For example, we will not show the first two kinds of scores for the \emph{test} set of PASCAL VOC 2012, since only the mean IoU is available.

\textbf{PASCAL VOC 2012.}
This dataset consists of photos taken in human daily life.
Besides the background category, there are twenty semantic categories, including bus, car, cat, sofa, monitor, etc.
There are 1,464 fully labeled images in the \emph{train} set and another 1,449 in the \emph{val} set.
Ground-truth labels of the 1,456 images in the \emph{test} set are not public, but there is an online evaluation server.
Following the conventional settings in the literature~\cite{FCN.CVPR.2015.Long,DeepLab.ICLR.2015.Chen}, we augment the \emph{train} set with extra labeled PASCAL VOC images from the semantic boundaries dataset~\cite{SBD.ICCV.2011.Hariharan}.
So, in total there are 10,582 images in the augmented \emph{train} set.

According to results on the \emph{val} set in~Table~\ref{tbl:fcrn voc val},
we notice the below three points.
1) Increasing the depth from 50 to 101 brings a significant improvement.
However, we observe no further improvement when increasing the depth to 152, probably, due to over-fitting.
2) Increasing the resolution of feature maps is beneficial.
3) Increasing the size of field-of-view~(FoV) to more than 224 is also helpful.
Note that 224 is the setting used in FCNs~\cite{FCN.CVPR.2015.Long}.

The above results suggest
using deeper networks, generating high resolution feature maps, and enlarging the FoV of classifiers.
This  makes sense because we can obtain richer and finer features, and  classifiers learn from larger context regions.
However, it also makes computational cost heavier and may be limited by the amount of GPU memories.
Besides, as for the size of FoV, there is another important factor to consider.
Note that all of the images in PASCAL VOC are no larger than 500$\times$500, and we feed a network with original images (without resizing) during testing.
Thus, we have to limit the size of FoV below 500 pixels on this dataset.
Otherwise, the dilated kernels of a classifier will be larger than the size of feature maps.
As a result, the outer part of the kernels must be applied to padded zeros, which may cause adversarial impact.
Similarly, if the size of FoV is larger than the size of image crops during training, the outer part of the kernels cannot be properly learned.
In~Table~\ref{tbl:fcrn voc val}, our largest evaluated size of FoV is 392.
No matter what is the depth, networks with this setting always achieve the best performance.
To realize such a large FoV, we can either enlarge the kernel size of the classifier or increase the dilation of these kernels.
However, this dilation should not be too large, since the feature vector per location can only cover a limited size of area.
For example, models with a dilation of eighteen show no advantages over those with a dilation of twelve.

\begin{table}[t]
\caption{Results of our vanilla FCRNs on the \emph{val} set of PASCAL VOC 2012.}
\label{tbl:fcrn voc val}
\centering
\small
\begin{tabular}{ccccc|ccc}
\toprule
Depth & Resolution & Kernel & Dilation & FoV & Pixel acc.~\% & Mean acc.~\% & Mean IoU~\% \\
\hline\hline
50 & 1/16 & 3 & 6 & 208 & 92.74 & 78.68 & 69.09 \\
50 & 1/8 & 3 & 6 & 104 & 92.50 & 77.60 & 67.61 \\
50 & 1/8 & 3 & 12 & 200 & 93.03 & 79.51 & 69.94 \\
50 & 1/8 & 3 & 18 & 296 & 93.02 & 79.28 & 70.01 \\
50 & 1/8 & 5 & 6 & 200 & 92.98 & 79.34 & 69.81 \\
50 & \textbf{1/8} & \textbf{5} & \textbf{12} & \textbf{392} & \textbf{93.25} & \textbf{79.84} & \textbf{71.10} \\
50 & 1/8 & 7 & 6 & 296 & 93.14 & 79.54 & 70.67 \\
\hline\hline
101 & 1/16 & 3 & 6 & 208 & 93.22 & 80.16 & 70.93 \\
101 & 1/8 & 3 & 6 & 104 & 93.20 & 79.87 & 70.20 \\
101 & 1/8 & 3 & 12 & 200 & 93.68 & 81.29 & 72.34 \\
101 & 1/8 & 3 & 18 & 296 & 93.67 & 81.15 & 72.37 \\
101 & 1/8 & 5 & 6 & 200 & 93.52 & 81.00 & 71.97 \\
101 & \textbf{1/8} & \textbf{5} & \textbf{12} & \textbf{392} & \textbf{93.87} & \textbf{81.87} & \textbf{73.41} \\
101 & 1/8 & 7 & 6 & 296 & 93.61 & 81.34 & 72.56 \\
\bottomrule
\end{tabular}
\end{table}

We compare our method with the previous best performers on the \emph{test} set in the bottom part of Table~\ref{tbl:comparison voc}.
Being trained with only the augmented PASCAL VOC data, our model outperforms the previous best performer by 3.8\% and wins the first place for 18 out of the 20 object categories.
Generally speaking, our method usually loses for those very hard categories, e.g., `bicycle' and `chair', for which most of the methods can only achieve scores below 60.0\%.
Instances of these categories are usually of large diversity and in occluded situations.
More importantly, some of them are given with quite elaborated annotations, e.g., a bicycle with carefully labeled skeletons.
The above facts suggest that more training data are required for these categories.
But unfortunately, this is not the case for this dataset.
Some works~\cite{CRFasRNN.ICCV.2015.Zheng,DPN.ICCV.2015.Liu,AdelaideContext.2016.Lin} pre-trained their models with the Microsoft COCO~\cite{COCO.ECCV.2014.Lin} data, which is composed of about 120k labeled images.
In this case, the best previous result is 77.8\%~\cite{AdelaideContext.2016.Lin}.
With this setting, we can only observe a rather limited improvement.
It seems like that more efforts should be put in domain adaption from COCO to PASCAL VOC.
Here, we leave this problem as one of our future works, and focus on the augmented PASCAL VOC dataset.
Nevertheless, even with less training data, our method still beats the previous best performer by 1.3\%.

\begin{savenotes}
\begin{table}[t]
\caption{Comparison of category-wise and mean IoU scores on the \emph{test} set of PASCAL VOC 2012.}
\vspace{-2.0mm}
\label{tbl:comparison voc}
\setlength{\tabcolsep}{1pt}
\centering
\footnotesize
\resizebox{0.9\textwidth}{!}
{
\begin{tabular}{r|cccccccccccccccccccc|c}
  \toprule
{Method} & \rot{aeroplane} & \rot{bicycle} & \rot{bird} & \rot{boat} & \rot{bottle} & \rot{bus} & \rot{car} & \rot{cat} & \rot{chair} & \rot{cow} & \rot{diningtable} & \rot{dog} & \rot{horse} & \rot{motorbike} & \rot{person} & \rot{pottedplant} & \rot{sheep} & \rot{sofa} & \rot{train} & \rot{tvmonitor} & \rot{Mean}   \\
\hline\hline
\multicolumn{22}{c}{Results on the \emph{val} set} \\
\hline
FCRN & 86.7 & 39.5 & 85.5 & 66.9 & 79.3 & 90.7 & 84.7 & 90.6 & 34.0 & 79.1 & 51.6 & 83.9 & 80.6 & 80.0 & 83.0 & 55.7 & 80.6 & 40.3 & 82.7 & 72.9 & 73.4 \\
\hline
FCRN + Bs. & 88.3 & 40.4 & 86.5 & 66.6 & 80.1 & 91.6 & 84.3 & 90.1 & 36.6 & 83.7 & 53.6 & 84.5 & 85.1 & 79.9 & 83.9 & 59.0 & 83.3 & 44.6 & 81.1 & 74.5 & 74.8 \\
\hline\hline
\multicolumn{22}{c}{Results on the \emph{test} set} \\
\hline
FCN-8s~\cite{FCN.CVPR.2015.Long} & 76.8 & 34.2 & 68.9 & 49.4 & 60.3 & 75.3 & 74.7 & 77.6 & 21.4 & 62.5 & 46.8 & 71.8 & 63.9 & 76.5 & 73.9 & 45.2 & 72.4 & 37.4 & 70.9 & 55.1 & 62.2 \\
DeepLab~\cite{DeepLab.ICLR.2015.Chen} & 84.4 & 54.5 & 81.5 & 63.6 & 65.9 & 85.1 & 79.1 & 83.4 & 30.7 & 74.1 & 59.8 & 79.0 & 76.1 & 83.2 & 80.8 & 59.7 & 82.2 & 50.4 & 73.1 & 63.7 & 71.6 \\
CRFasRNN~\cite{CRFasRNN.ICCV.2015.Zheng} & 87.5 & 39.0 & 79.7 & 64.2 & 68.3 & 87.6 & 80.8 & 84.4 & 30.4 & 78.2 & 60.4 & 80.5 & 77.8 & 83.1 & 80.6 & 59.5 & 82.8 & 47.8 & 78.3 & 67.1 & 72.0 \\
DeconvNet~\cite{DeconvNet.ICCV.2015.Noh} & 89.9 & 39.3 & 79.7 & 63.9 & 68.2 & 87.4 & 81.2 & 86.1 & 28.5 & 77.0 & 62.0 & 79.0 & 80.3 & 83.6 & 80.2 & 58.8 & 83.4 & 54.3 & 80.7 & 65.0 & 72.5 \\
DPN~\cite{DPN.ICCV.2015.Liu} & 87.7 & \textbf{59.4} & 78.4 & 64.9 & 70.3 & 89.3 & 83.5 & 86.1 & 31.7 & 79.9 & 62.6 & 81.9 & 80.0 & 83.5 & 82.3 & 60.5 & 83.2 & 53.4 & 77.9 & 65.0 & 74.1 \\
UoAContext~\cite{AdelaideContext.2016.Lin} & 90.6 & 37.6 & 80.0 & 67.8 & 74.4 & 92.0 & 85.2 & 86.2 & \textbf{39.1} & 81.2 & 58.9 & 83.8 & 83.9 & 84.3 & 84.8 & 62.1 & 83.2 & 58.2 & 80.8 & 72.3 & 75.3 \\
\hline
ours & \textbf{91.9} & 48.1 & \textbf{93.4} & \textbf{69.3} & \textbf{75.5} & \textbf{94.2} & \textbf{87.5} & \textbf{92.8} & 36.7 & \textbf{86.9} & \textbf{65.2} & \textbf{89.1} & \textbf{90.2} & \textbf{86.5} & \textbf{87.2} & \textbf{64.6} & \textbf{90.1} & \textbf{59.7} & \textbf{85.5} & \textbf{72.7} & \textbf{79.1} \\
\bottomrule
\end{tabular}
}
\vspace{-2.0mm}
\end{table}
\end{savenotes}

\textbf{Cityscapes.}
This dataset consists of street scene images taken using car-carried cameras.
There are nineteen semantic categories, including road, car, pedestrian, bicycle, etc.
There are 2975 fully labeled images in the \emph{train} set and another 500 in the \emph{val} set.
Ground-truth labels of images in the \emph{test} set are not public, but there is an online evaluation server also.
All of the images in this dataset are in the same size.
They are 1024 pixels high and 2048 pixels wide.

We show results on the \emph{val} set of Cityscapes in~Table~\ref{tbl:fcrn cityscapes val}.
Most of the observations on this dataset are consistent with those on PASCAL VOC 2012, as demonstrated previously.
Two notable exceptions are as follows.
First, the problem of over-fitting seems less severe.
One possible reason is that the resolution of images in this dataset are higher than those in PASCAL VOC 2012, so the total number of pixels are actually larger.
On the other hand, the diversity of images in this dataset is smaller than those in PASCAL VOC 2012.
In this sense, even less training data can cover a larger proportion of possible situations, which can reduce over-fitting.
Second, 392 seems still smaller than the optimal size of FoV.
Since the original images are in a size of 1024$\times$2048, we can feed a 50-layer network with larger image crops during both training and testing.
In this case, a network may prefer even larger FoV.
Therefore, to some extent, the ideal size of FoV depends on the size of image crops during training and testing.

\begin{table}[t]
\caption{Results of our vanilla FCRNs on the \emph{val} set of Cityscapes.}
\label{tbl:fcrn cityscapes val}
\centering
\small
\begin{tabular}{ccccc|ccc}
\toprule
Depth & Resolution & Kernel & Dilation & FoV & Pixel acc.~\% & Mean acc.~\% & Mean IoU~\% \\
\hline\hline
50 & 1/16 & 3 & 6 & 208 & 93.83 & 74.67 & 66.41 \\
50 & 1/8 & 3 & 6 & 104 & 94.38 & 74.89 & 66.58 \\
50 & 1/8 & 3 & 12 & 200 & 94.47 & 75.91 & 67.68 \\
50 & 1/8 & 3 & 18 & 296 & 94.53 & 76.52 & 68.38 \\
50 & 1/8 & 5 & 6 & 200 & 94.48 & 76.17 & 68.04 \\
50 & 1/8 & 5 & 12 & 392 & 94.61 & 76.68 & 68.71 \\
50 & 1/8 & 5 & 18 & 584 & \textbf{94.64} & 76.34 & 68.53 \\
50 & 1/8 & 7 & 6 & 296 & 94.58 & \textbf{76.88} & \textbf{68.79} \\
50 & 1/8 & 7 & 12 & 584 & \textbf{94.64} & 76.57 & \textbf{68.79} \\
\hline\hline
101 & 1/16 & 3 & 6 & 208 & 94.11 & 76.26 & 67.62 \\
101 & 1/8 & 3 & 6 & 104 & 94.68 & 77.15 & 68.58 \\
101 & 1/8 & 3 & 12 & 200 & 94.78 & 78.30 & 69.99 \\
101 & 1/8 & 3 & 18 & 296 & 94.82 & 78.21 & 70.00 \\
101 & 1/8 & 5 & 6 & 200 & 94.75 & 78.11 & 69.89 \\
101 & \textbf{1/8} & \textbf{5} & \textbf{12} & \textbf{392} & \textbf{94.87} & \textbf{79.17} & \textbf{71.16} \\
101 & 1/8 & 7 & 6 & 296 & 94.75 & 78.43 & 70.40 \\
\hline\hline
152 & 1/16 & 3 & 6 & 208 & 94.26 & 76.89 & 68.30 \\
152 & 1/8 & 3 & 6 & 104 & 94.82 & 78.30 & 69.69 \\
152 & 1/8 & 3 & 12 & 200 & 94.94 & 78.79 & 70.66 \\
152 & 1/8 & 3 & 18 & 296 & 94.93 & 79.19 & 70.92 \\
152 & 1/8 & 5 & 6 & 200 & 94.88 & 78.77 & 70.61 \\
152 & \textbf{1/8} & \textbf{5} & \textbf{12} & \textbf{392} & \textbf{95.00} & \textbf{79.38} & \textbf{71.51} \\
152 & 1/8 & 7 & 6 & 296 & 94.91 & 79.08 & 70.87 \\
\bottomrule
\end{tabular}
\end{table}

Our best model achieves an IoU score of 74.6\% on the \emph{val} set, as shown in~Table~\ref{tbl:bootstrapping},
which is compared with the previously reported best result 68.6\%~\cite{AdelaideContext.2016.Lin}.

\begin{savenotes}
\begin{table}[t]
\caption{Comparison of category-wise and mean IoU scores on the \emph{val} set of Cityscapes.}
\label{tbl:comparison cityscapes}
\setlength{\tabcolsep}{1pt}
\centering
\resizebox{0.95\textwidth}{!}
{
\begin{tabular}{r|ccccccccccccccccccc|c}
\toprule
{Method} & \rot{road} & \rot{sidewalk} & \rot{building} & \rot{wall} & \rot{fence} & \rot{pole} & \rot{traffic light} & \rot{traffic sign} & \rot{vegetation} & \rot{terrain} & \rot{sky} & \rot{person} & \rot{rider} & \rot{car} & \rot{truck} & \rot{bus} & \rot{train} & \rot{motorcycle} & \rot{bicycle} & \rot{Mean}
\\
\hline\hline
\multicolumn{21}{c}{Results on val set} \\
\hline
FCRN & 97.4 & 80.3 & 90.8 & 47.6 & 53.8 & 53.1 & 58.1 & 70.2 & 91.2 & 59.6 & 93.2 & 77.1 & 54.4 & 93.0 & 67.1 & 79.4 & 62.2 & 57.3 & 72.7 & 71.5 \\
\hline
FCRN+Bs. & 97.6 & 82.0 & 91.7 & 52.3 & 56.2 & 57.0 & 65.7 & 74.4 & 91.7 & 62.5 & 93.8 & 79.8 & 59.6 & 94.0 & 66.2 & 83.7 & 70.3 & 64.2 & 75.5 & 74.6 \\
\bottomrule
\end{tabular}
}
\end{table}
\end{savenotes}

\textbf{PASCAL-Context.}
This dataset is composed of PASCAL VOC 2010 images with extra object and \emph{stuff} labels, e.g., bag, food, sign, ceiling, ground and snow.
Including the background category, in total there are sixty semantic categories.
At present, labels on the \emph{test} set are not released yet.
So, we train networks using the 4,998 images in the \emph{train} set, and evaluate them using the 5,105 images in the \emph{val} set.
For this dataset, we just evaluate the same model as the one used for PASCAL VOC 2012.
Our method outperforms the previous best performers with a clear margin in terms of all the three kinds of scores, as shown in~Table~\ref{tbl:comparison pascal-context}.

\begin{table}[t]
\caption{Comparison on the \emph{val} set of PASCAL-Context.}
\label{tbl:comparison pascal-context}
\centering
\resizebox{0.7\textwidth}{!}
{
\begin{tabular}{r|c|c|c}
\toprule
\multicolumn{1}{c|}{Method} & Pixel acc.~\% & Mean acc.~\% & Mean IoU~\% \\
\hline
FCN-8s~\cite{FCN.CVPR.2015.Long} & 65.9 & 46.5 & 35.1 \\
BoxSup~\cite{BoxSup.ICCV.2015.Dai} & -- & -- & 40.5 \\
UoA-Context~\cite{AdelaideContext.2016.Lin} & 71.5 & 53.9 & 43.3 \\
\hline\
ours & \textbf{72.9} & \textbf{54.8} & \textbf{44.5} \\
\bottomrule
\end{tabular}
}
\end{table}

\subsection{Instance-level segmentation results}
We now evaluate the whole framework in terms of instance segmentation on the PASCAL VOC 2012 dataset, following the common protocols used in several recent works~\cite{SDS.ECCV.2014.Hariharan,HyperColumn.CVPR.2015.Hariharan,MNC.CVPR.2016.Dai}.
We use the annotations in SBD~\cite{SBD.ICCV.2011.Hariharan}.
According to the PASCAL VOC 2012 splits, there are 5,623 images in the \emph{train} set and 5,732 in the \emph{val} set.
Since there is no annotation for the \emph{test} set, we train models with the \emph{train} set and evaluate them with the \emph{val} set.
We report two mean region average precisions $\textrm{mAP}^{r}_{0.5}$ and $\textrm{mAP}^{r}_{0.7}$,
and also the mean volume region average precisions $\textrm{mAP}^{r}_{vol}$,
which were proposed by Hariharan~et~al.~\cite{SDS.ECCV.2014.Hariharan}.

We show results on the \emph{val} set of PASCAL VOC 2012 in~Table~\ref{tbl:comparison voc instance}.
Our method can finally achieve \emph{on par} or even better performance compared with the previous best performers.
Especially, in terms of the mean AP at an overlap of 0.7, our method outperforms the previous best one by 3.1\%, which is a significant improvement.
By pre-training the semantic segmentation network using the COCO dataset~\cite{COCO.ECCV.2014.Lin},
we can further improve the performance by 2.0\%.
In a pilot experiment, we generate semantic score maps with ground-truth masks,
while still compute transform maps with our best localization network.
The performance goes up to 73.0\% in $\textrm{mAP}^{r}_{0.5}$ and 60.6\% in $\textrm{mAP}^{r}_{0.7}$,
which shows the great potential of our method to benefit from the improvement of semantic category-level segmentation approaches.

%
%
%

%
%
%
%
%
%

\begin{table}[t]
\caption{Comparison on the \emph{val} set of PASCAL VOC 2012. Bs.~means `bootstrapping'.}
\vspace{-2.0mm}
\label{tbl:comparison voc instance}
\footnotesize
\centering
{
\begin{tabular}{l|c|c|c}
\toprule
\multicolumn{1}{c|}{Method} & $\textrm{mAP}^\textit{$r$}_{0.5}$~\% & $\textrm{mAP}^\textrm{$r$}_{0.7}$~\% & $\textrm{mAP}^\textrm{$r$}_{vol}$~\% \\
\hline
SDS~\cite{SDS.ECCV.2014.Hariharan} & 49.7 & 25.3 & 41.4 \\
Hypercolumn~\cite{HyperColumn.CVPR.2015.Hariharan} & 60.0 & 40.4 & -- \\
MNC~\cite{MNC.CVPR.2016.Dai} & \textbf{63.5} & 41.5 & -- \\
\hline
\phantom{0}50-layer & 57.2 & 40.5 & 52.5 \\
\phantom{0}50-layer, Bs. & 58.7 & 42.0 & 53.8 \\
\phantom{0}50-layer, Bs., weighted loss & 59.8 & 43.5 & 54.7 \\
101-layer, Bs., weighted loss & 60.9 & 44.6 & 55.5 \\
101-layer, Bs., weighted loss, COCO & 61.5 & \textbf{46.6} & {\bf 56.4} \\
\bottomrule
\end{tabular}
}
\end{table}

\subsection{Importance of online bootstrapping of hard training pixels}
We show the results of two best models for semantic segmentation in~Table~\ref{tbl:bootstrapping}.
In both cases, the best setting is to keep the 512 most hardest pixels.
Particularly, the proposed bootstrapping improves the mean IoU by 3.1\% on Cityscapes, which is a  significant margin.
The problem of biased data is more severe on this dataset, since there are clearly much more `sky' and `road' pixels than `traffic sign' in a street scene.
We also show the category-wise results in~Tables~\ref{tbl:comparison voc}~and~\ref{tbl:comparison cityscapes}.
Generally, the proposed bootstrapping can clearly improve the performance for those categories which are less frequent in training data, e.g., `cow' and `horse' on PASCAL VOC 2012, `traffic light' and `train' on Cityscapes.
Similarly, the proposed online bootstrapping also contributes clearly to our localization networks, as shown in~Table~\ref{tbl:comparison voc instance}.

\begin{table}[h]
\caption{Results showing the impact of online bootstrapping.}
\label{tbl:bootstrapping}
\centering
\small
\resizebox{0.95\textwidth}{!}
{
\begin{tabular}{ccccc|ccc}
\toprule
Depth & Resolution & Kernel & Dilation & Bs. & Pixel acc.~\% & Mean acc.~\% & Mean IoU~\% \\
\hline\hline
\multicolumn{8}{c}{PASCAL VOC 2012} \\
\hline\hline
101 & 1/8 & 5 & 12 & F & 93.87 & 81.87 & 73.41 \\
\hline
101 & 1/8 & 5 & 12 & 256 & 94.11 & 81.44 & 74.41 \\
101 & 1/8 & 5 & 12 & \textbf{512} & \textbf{94.23} & \textbf{82.09} & \textbf{74.80} \\
101 & 1/8 & 5 & 12 & 1024 & 94.08 & 81.84 & 74.17 \\
\hline
\multicolumn{8}{c}{Cityscapes} \\
\hline\hline
152 & 1/8 & 5 & 12 & F & 95.00 & 79.38 & 71.51 \\
\hline
152 & 1/8 & 5 & 12 & 256 & 95.41 & 81.37 & 73.97 \\
152 & 1/8 & 5 & 12 & \textbf{512} & \textbf{95.46} & \textbf{82.04} & \textbf{74.64} \\
152 & 1/8 & 5 & 12 & 1024 & 95.38 & 81.00 & 73.45 \\
\bottomrule
\end{tabular}
}
\end{table}

\subsection{Qualitative results}

\begin{figure}[t!]
\centering
\includegraphics[width=0.98\linewidth,trim=0 150 280 0]{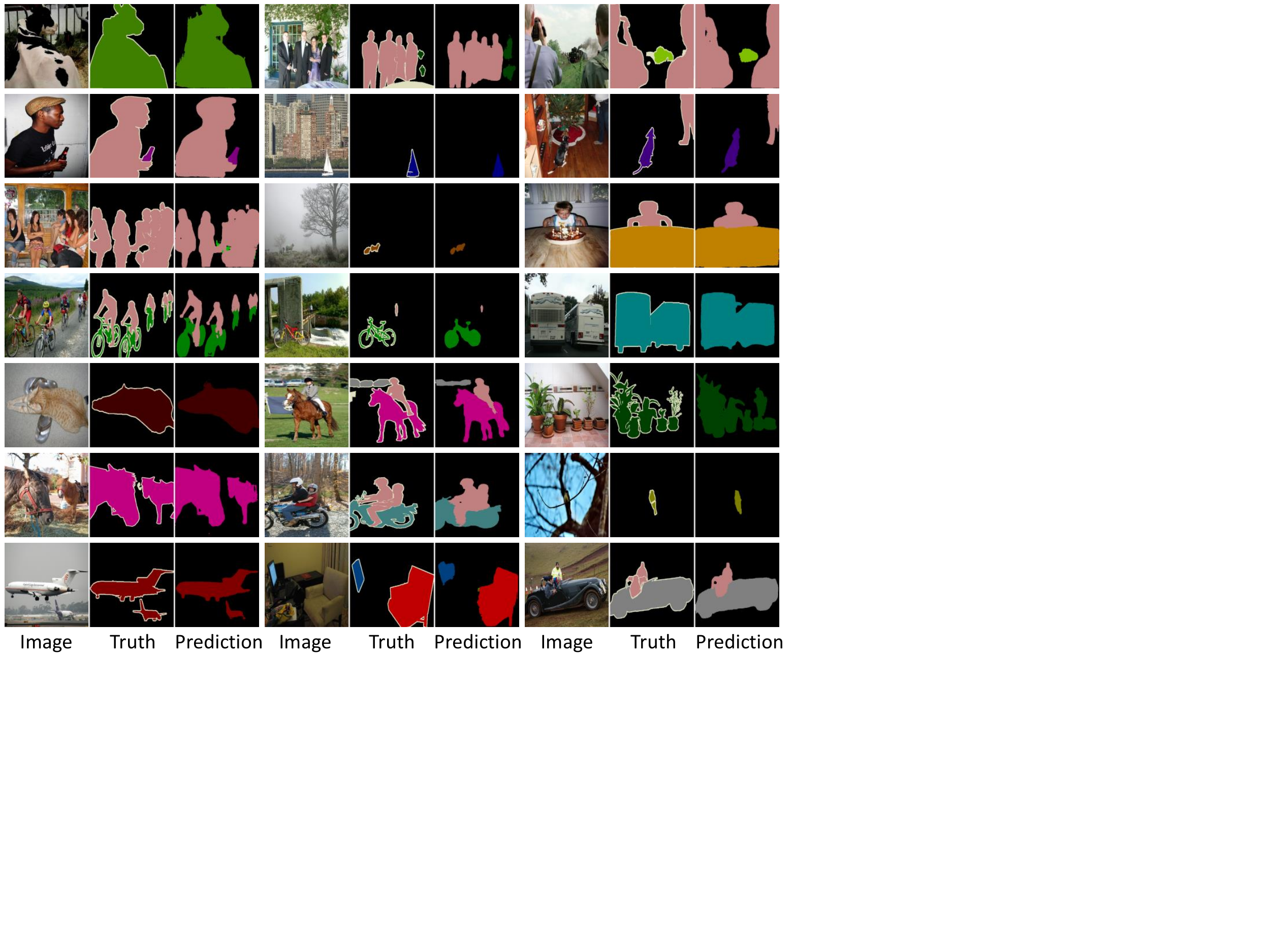}
\caption{The qualitative results of our method on PASCAL VOC 2012 for semantic segmentation.}
\label{fig:voc_qualitative}
\end{figure}

\begin{figure}[t!]
\centering
\includegraphics[width=0.98\linewidth,trim=0 290 280 0]{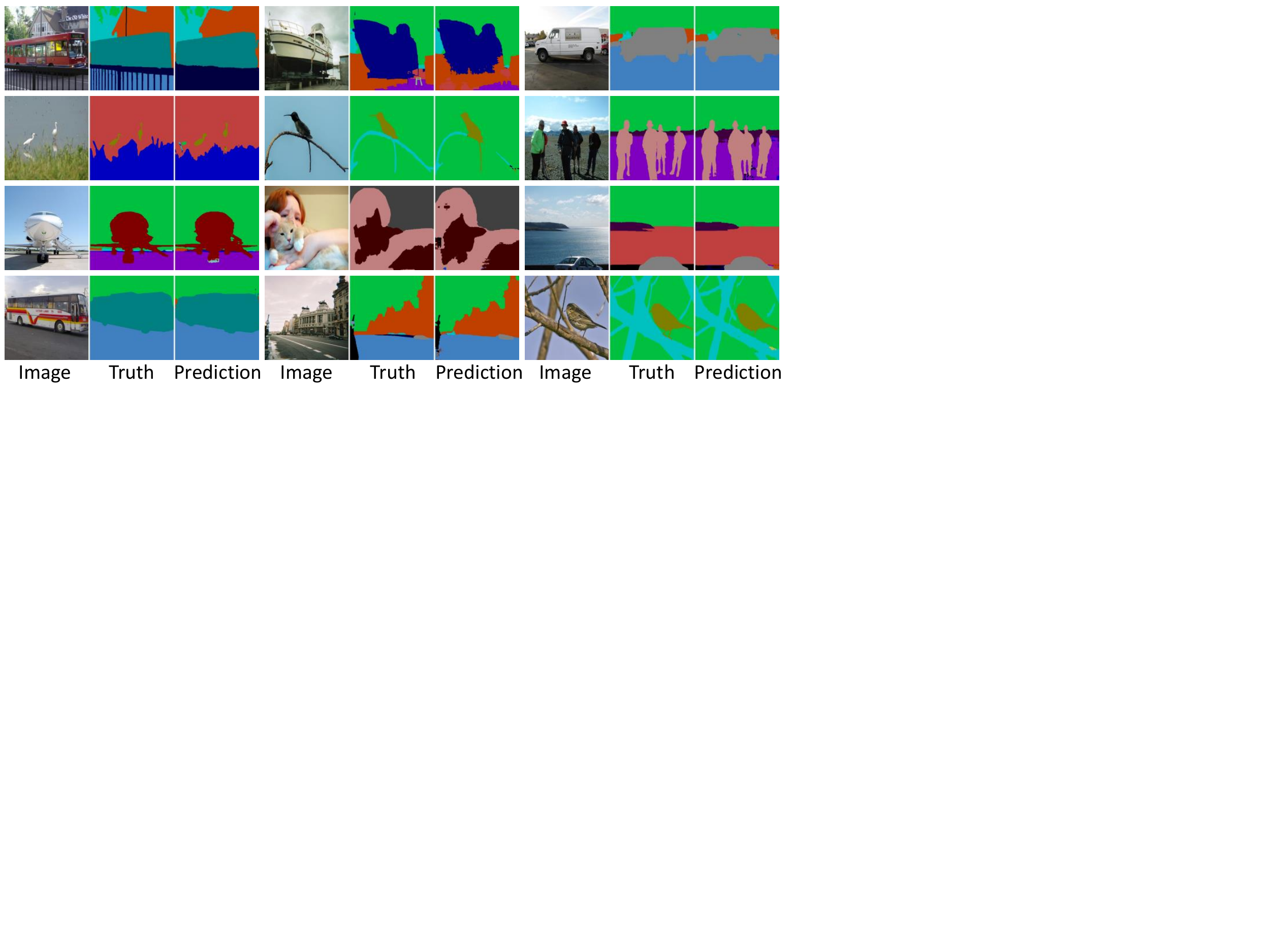}
\caption{The qualitative results of our method on PASCAL-Context for semantic segmentation.}
\label{fig:pascal-context_qualitative}
\end{figure}

\begin{figure}[t!]
\centering
\includegraphics[width=0.98\linewidth,trim=0 220 280 0]{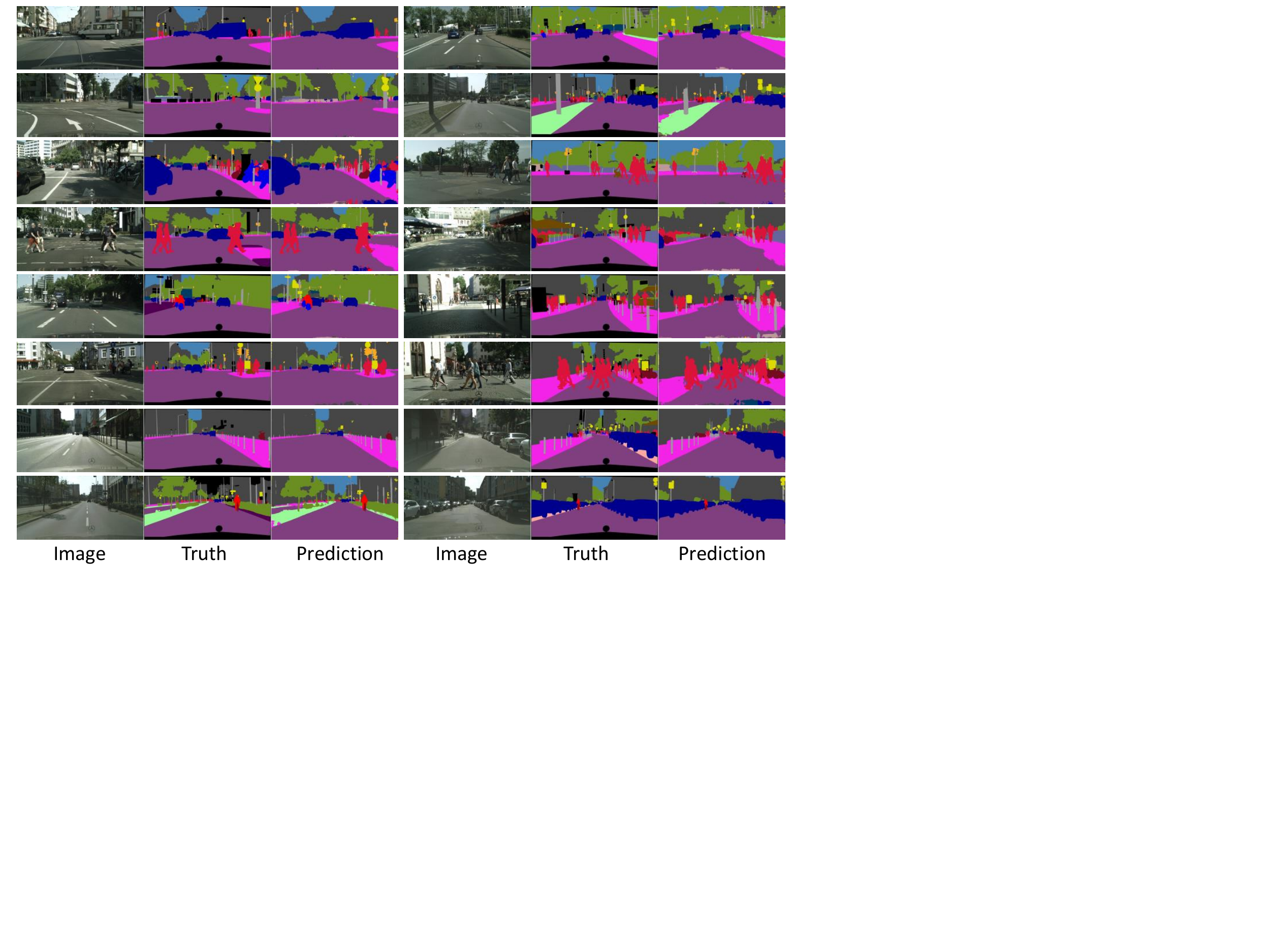}
\caption{The qualitative results of our method on Cityscapes for semantic segmentation.}
\label{fig:cityscapes_qualitative}
\end{figure}

\begin{figure}[t!]
\centering
\includegraphics[width=0.8\linewidth,trim=0 0 120 0]{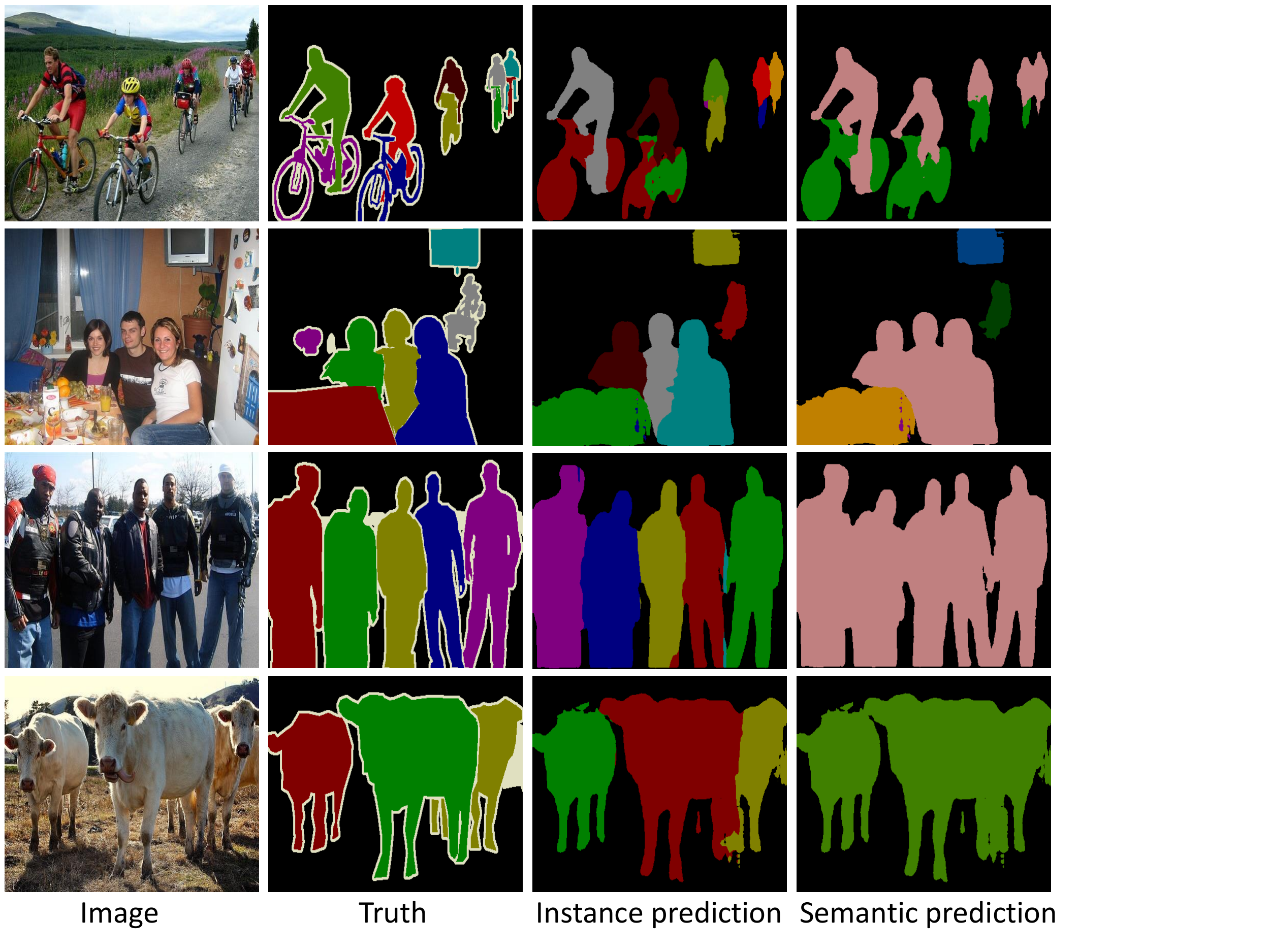}
\caption{The qualitative results of our method on PASCAL VOC 2012 for instance segmentation.}
\label{fig:voc_instance_qualitative}
\end{figure}

We show qualitative results for semantic segmentation in~Figs.~\ref{fig:voc_qualitative}~,~\ref{fig:pascal-context_qualitative}~and~\ref{fig:cityscapes_qualitative}, as well as for instance segmentation in~Figs.~\ref{fig:voc_instance_qualitative}.
Note that the white borders and regions in the ground-truth in~Figs.~\ref{fig:voc_qualitative}~and~\ref{fig:voc_instance_qualitative} are excluded during evaluation,
as well as those black regions in the ground-truth in~Fig.~\ref{fig:cityscapes_qualitative}.

\section{Conclusions}

In this work, we have built a state-of-the-art fully convolutional residual network for semantic category-level
segmentation.
 On top of this model,
we have proposed a new pipeline for instance-level segmentation,
which is intrinsically different from  the currently commonly-used bounding-box detection based methods.
We have also proposed an online bootstrapping approach, which contributes greatly in both of the above two tasks.
Finally, on the PASCAL VOC 2012 dataset, we have achieved the currently best mean IoU score for semantic segmentation,
and the state-of-the-art performance for instance-level segmentation.

\newpage

\small
\newpage
\bibliographystyle{IEEEtran}
\bibliography{CSRef}

\begin{thebibliography}{10}
\providecommand{\url}[1]{#1}
\csname url@samestyle\endcsname
\providecommand{\newblock}{\relax}
\providecommand{\bibinfo}[2]{#2}
\providecommand{\BIBentrySTDinterwordspacing}{\spaceskip=0pt\relax}
\providecommand{\BIBentryALTinterwordstretchfactor}{4}
\providecommand{\BIBentryALTinterwordspacing}{\spaceskip=\fontdimen2\font plus
\BIBentryALTinterwordstretchfactor\fontdimen3\font minus
  \fontdimen4\font\relax}
\providecommand{\BIBforeignlanguage}[2]{{%
\expandafter\ifx\csname l@#1\endcsname\relax
\typeout{** WARNING: IEEEtran.bst: No hyphenation pattern has been}%
\typeout{** loaded for the language `#1'. Using the pattern for}%
\typeout{** the default language instead.}%
\else
\language=\csname l@#1\endcsname
\fi
#2}}
\providecommand{\BIBdecl}{\relax}
\BIBdecl

\bibitem{FCN.CVPR.2015.Long}
J.~Long, E.~Shelhamer, and T.~Darrell, ``Fully convolutional networks for
  semantic segmentation,'' in \emph{Proc. IEEE Conf. Comp. Vis. Patt. Recogn.},
  2015.

\bibitem{DeepLab.ICLR.2015.Chen}
L.~Chen, G.~Papandreou, I.~Kokkinos, K.~Murphy, and A.~Yuille, ``Semantic image
  segmentation with deep convolutional nets and fully connected {CRF}s,'' in
  \emph{Proc. Int. Conf. Learn. Representations}, 2015.

\bibitem{AdelaideContext.2016.Lin}
G.~Lin, C.~Shen, A.~van~den Hengel, and I.~Reid, ``Exploring context with deep
  structured models for semantic segmentation,'' arXiv:1603.03183, 2016.

\bibitem{ILSVRC2015}
O.~Russakovsky, J.~Deng, H.~Su, J.~Krause, S.~Satheesh, S.~Ma, Z.~Huang,
  A.~Karpathy, A.~Khosla, M.~Bernstein, A.~Berg, and L.~Fei-Fei, ``Image{N}et
  {L}arge {S}cale {V}isual {R}cognition {C}hallenge,'' 2015.

\bibitem{VGGNet.2014.Simonyan}
K.~Simonyan and A.~Zisserman, ``Very deep convolutional networks for
  large-scale image recognition,'' arXiv:1409.1556, 2014.

\bibitem{ResNet.CVPR.2016.He}
K.~He, X.~Zhang, S.~Ren, and J.~Sun, ``Deep residual learning for image
  recognition,'' in \emph{Proc. IEEE Conf. Comp. Vis. Patt. Recogn.}, 2016.

\bibitem{ResNet2.2016.He}
------, ``Identity mappings in deep residual networks,'' arXiv:1603.05027,
  2016.

\bibitem{SDS.ECCV.2014.Hariharan}
B.~Hariharan, P.~Arbel{\'{a}}ez, R.~Girshick, and J.~Malik, ``Simultaneously
  detection and segmentation,'' in \emph{Proc. Eur. Conf. Comp. Vis.}, 2014.

\bibitem{HyperColumn.CVPR.2015.Hariharan}
------, ``Hypercolumns for object segmentation and fine-grained localization,''
  in \emph{Proc. IEEE Conf. Comp. Vis. Patt. Recogn.}, 2015.

\bibitem{MNC.CVPR.2016.Dai}
J.~Dai, K.~He, and J.~Sun, ``Instance-aware semantic segmentation via
  multi-task network cascades,'' in \emph{Proc. IEEE Conf. Comp. Vis. Patt.
  Recogn.}, 2016.

\bibitem{FastRCNN.ICCV.2015.Girshick}
R.~Girshick, ``Fast {R}-{CNN},'' in \emph{Proc. IEEE Int. Conf. Comp. Vis.},
  2015.

\bibitem{ProposalFree.2015.Liang}
X.~Liang, Y.~Wei, X.~Shen, J.~Yang, L.~Lin, and S.~Yan, ``Proposal-free network
  for instance-level object segmentation,'' arXiv:1509.02636, 2015.

\bibitem{CityscapesInstance.2016.Uhrig}
J.~Uhrig, M.~Cordts, U.~Franke, and T.~Brox, ``Pixel-level encoding and depth
  layering for instance-level semantic labeling,'' arXiv:1604.05096, 2016.

\bibitem{CNN.NIPS.2012.Krizhevsky}
A.~Krizhevsky, I.~Sutskever, and G.~Hinton, ``Image{N}et classification with
  deep convolutional neural networks,'' in \emph{Proc. Advances in Neural Inf.
  Process. Syst.}, 2012.

\bibitem{GoogLeNet.2014.Szegedy}
C.~Szegedy, W.~Liu, Y.~Jia, P.~Sermanet, S.~Reed, D.~Anguelov, D.~Erhan,
  V.~Vanhoucke, and A.~Robinovich, ``Going deeper with convolutions,''
  arXiv:1409.4842, 2014.

\bibitem{Highway}
R.~K. Srivastava, K.~Greff, and J.~Schmidhuber, ``Training very deep
  networks,'' in \emph{Proc. Advances in Neural Inf. Process. Syst.}, 2015.

\bibitem{DenseCRF.NIPS.2011.Krahenbuhl}
P.~Kr{\"{a}}henb{\"{u}}hl and V.~Koltun, ``Efficient inference in fully
  connected {CRF}s with gaussian edge potentials,'' in \emph{Proc. Advances in
  Neural Inf. Process. Syst.}, 2011.

\bibitem{CRFasRNN.ICCV.2015.Zheng}
S.~Zheng, S.~Jayasumana, B.~Romera-Paredes, V.~Vineet, Z.~Su, D.~Du, C.~Huang,
  and P.~Torr, ``Conditional random fields as recurrent neural networks,'' in
  \emph{Proc. IEEE Int. Conf. Comp. Vis.}, 2015.

\bibitem{ISM.IJCV.2007.Leibe}
B.~Leibe, A.~Leonardis, and B.~Schiele, ``Robust object detection with
  interleaved categorization and segmentation,'' \emph{Int. J. Comput. Vision},
  2007.

\bibitem{OHEM.ICLR.2016.Loshchilov}
I.~Loshchilov and F.~Hutter, ``Online batch selection for faster training of
  neural networks,'' in \emph{Proc. Int. Conf. Learn. Representations}, 2016.

\bibitem{DetOHEM.CVPR.2016.Shrivastava}
A.~Shrivastava, A.~Gupta, and R.~Girshick, ``Training region-based object
  detectors with online hard example mining,'' in \emph{Proc. IEEE Conf. Comp.
  Vis. Patt. Recogn.}, 2016.

\bibitem{WaveletTour.2008.Mallat}
S.~Mallat, \emph{A wavelet tour of signal processing}, 3rd~ed.\hskip 1em plus
  0.5em minus 0.4em\relax Academic Press, December 2008.

\bibitem{Caffe.2014.Jia}
Y.~Jia, E.~Shelhamer, J.~Donahue, S.~Karayev, J.~Long, R.~Girshick,
  S.~Guadarrama, and T.~Darrell, ``Caffe: {C}onvolutional architecture for fast
  feature embedding,'' arXiv:1408.5093, 2014.

\bibitem{PascalVoc.IJCV.2014.Everingham}
M.~Everingham, S.~Eslami, L.~van Gool, C.~Williams, J.~Winn, and A.~Zisserman,
  ``The {PASCAL} visual object classes challenge: {A} retrospective,''
  \emph{Int. J. Comput. Vision}, 2014.

\bibitem{Cityscapes.CVPR.2016.Cordts}
M.~Cordts, M.~Omran, S.~Ramos, T.~Rehfeld, M.~Enzweiler, R.~Benenson,
  U.~Franke, S.~Roth, and B.~Schiele, ``The {C}ityscapes dataset for semantic
  urban scene understanding,'' in \emph{Proc. IEEE Conf. Comp. Vis. Patt.
  Recogn.}, 2016.

\bibitem{PascalContext.CVPR.2014.Mottaghi}
R.~Mottaghi, X.~Chen, X.~Liu, N.~Cho, S.~Lee, S.~Fidler, R.~Urtasun, and
  A.~Yuille, ``The role of context for object detection and semantic
  segmentation in the wild,'' in \emph{Proc. IEEE Conf. Comp. Vis. Patt.
  Recogn.}, 2014.

\bibitem{SBD.ICCV.2011.Hariharan}
B.~Hariharan, P.~Arbel{\'{a}}ez, L.~Bourdev, S.~Maji, and J.~Malik, ``Semantic
  contours from inverse detectors,'' in \emph{Proc. IEEE Int. Conf. Comp.
  Vis.}, 2011.

\bibitem{DPN.ICCV.2015.Liu}
Z.~Liu, X.~Li, P.~Luo, C.~Loy, and X.~Tang, ``Semantic image segmentation via
  deep parsing network,'' in \emph{Proc. IEEE Int. Conf. Comp. Vis.}, 2015.

\bibitem{COCO.ECCV.2014.Lin}
T.~Lin, M.~Maire, S.~Belongie, J.~Hays, P.~Perona, D.~R.~P. Doll{\'{a}}r, and
  C.~Zitnick, ``Microsoft {COCO}: {C}ommon objects in context,'' in \emph{Proc.
  Eur. Conf. Comp. Vis.}, 2014.

\bibitem{DeconvNet.ICCV.2015.Noh}
H.~Noh, S.~Hong, and B.~Han, ``Learning deconvolution network for semantic
  segmentation,'' in \emph{Proc. IEEE Int. Conf. Comp. Vis.}, 2015.

\bibitem{BoxSup.ICCV.2015.Dai}
J.~Dai, K.~He, and J.~Sun, ``Box{S}up: {E}xploiting bounding boxes to supervise
  convolutional networks for semantic segmentation,'' in \emph{Proc. IEEE Int.
  Conf. Comp. Vis.}, 2015.

\end{thebibliography}

\end{document}